\documentclass[sigconf]{acmart}
\AtBeginDocument{%
  \providecommand\BibTeX{{%
    \normalfont B\kern-0.5em{\scshape i\kern-0.25em b}\kern-0.8em\TeX}}}

\setcopyright{acmcopyright}
\copyrightyear{2021} 
\acmYear{2021} 
\setcopyright{acmcopyright}\acmConference[MM '21]{Proceedings of the 29th ACM International Conference on Multimedia}{October 20--24, 2021}{Virtual Event, China}
\acmBooktitle{Proceedings of the 29th ACM International Conference on Multimedia (MM '21), October 20--24, 2021, Virtual Event, China}
\acmPrice{15.00}
\acmDOI{10.1145/3474085.3475412}
\acmISBN{978-1-4503-8651-7/21/10}

\acmSubmissionID{1410}

\newcommand{\myparagraph}[1]{\vspace{0.1em}\noindent\textbf{#1}}

\begin{document}
\fancyhead{}

\title{iButter: Neural Interactive Bullet Time Generator for Human Free-viewpoint Rendering}

\author{Liao Wang}
\affiliation{%
  \institution{Shanghaitech University}
  \city{Shanghai}
  \country{China}
}
\email{wangla@shanghaitech.edu.cn}

\author{Ziyu Wang}
\affiliation{%
  \institution{Shanghaitech University}
  \city{Shanghai}
  \country{China}
}
\email{wangzy6@shanghaitech.edu.cn}

\author{Pei Lin}
\affiliation{%
  \institution{Shanghaitech University}
  \city{Shanghai}
  \country{China}
}
\email{linpei@shanghaitech.edu.cn}

\author{Yuheng Jiang}
\affiliation{%
  \institution{Shanghaitech University}
  \city{Shanghai}
  \country{China}
}
\email{jiangyh2@shanghaitech.edu.cn}

\author{Xin Suo}
\affiliation{%
  \institution{Shanghaitech University}
  \city{Shanghai}
  \country{China}
}
\email{suoxin@shanghaitech.edu.cn}

\author{Minye Wu}
\affiliation{%
  \institution{Shanghaitech University}
  \city{Shanghai}
  \country{China}
}
\email{wumy@shanghaitech.edu.cn}

\author{Lan Xu}
\affiliation{%
  \institution{Shanghaitech University}
  \city{Shanghai}
  \country{China}
}
\email{xulan1@shanghaitech.edu.cn}

\author{Jingyi Yu}
\affiliation{%
  \institution{Shanghai Engineering Research
Center of Intelligent Vision and
Imaging, School of Information
Science and Technology,
ShanghaiTech University}
\city{Shanghai}
  \country{China}
}
\email{yujingyi@shanghaitech.edu.cn}

\begin{abstract}
Generating ``bullet-time'' effects of human free-viewpoint videos is critical for immersive visual effects and VR/AR experience.
Recent neural advances still lack the controllable and interactive bullet-time design ability for human free-viewpoint rendering, especially under the real-time, dynamic and general setting for our trajectory-aware task.
To fill this gap, in this paper we propose a neural interactive bullet-time generator (iButter) for photo-realistic human free-viewpoint rendering from dense RGB streams, which enables flexible and interactive design for human bullet-time visual effects.
Our iButter approach consists of a real-time preview and design stage as well as a trajectory-aware refinement stage.
During preview, we propose an interactive bullet-time design approach by extending the NeRF rendering to a real-time and dynamic setting and getting rid of the tedious per-scene training.
To this end, our bullet-time design stage utilizes a hybrid training set, light-weight network design and an efficient silhouette-based sampling strategy.
During refinement, we introduce an efficient trajectory-aware scheme within 20 minutes, which jointly encodes the spatial, temporal consistency and semantic cues along the designed trajectory, achieving photo-realistic bullet-time viewing experience of human activities.
Extensive experiments demonstrate the effectiveness of our approach for convenient interactive bullet-time design and photo-realistic human free-viewpoint video generation.

\end{abstract}



\begin{teaserfigure}
\vspace{-8pt}
  \includegraphics[width=\textwidth]{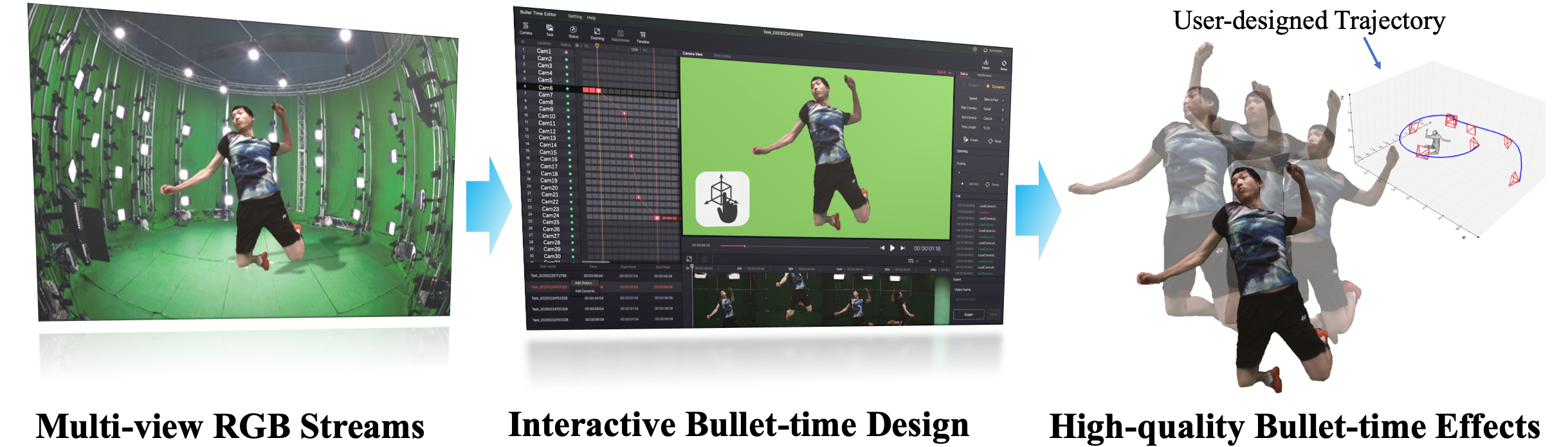}
  \vspace{-20pt}
  \caption{Our neural interactive bullet-time generator (iButter) enables convenient, flexible and interactive design for human bullet-time visual effects from dense RGB streams, and achieves high-quality and photo-realistic human performance rendering along the designed trajectory.}
  \label{fig:teaser}
\end{teaserfigure}

\maketitle

\section{Introduction}
Novel view synthesis has been widely used in visual effects to provide unique viewing experiences.
One of the most famous examples is the ``bullet-time'' effects presented in the feature film The Matrix, which creates
the stopping-time illusion with smooth transitions of viewpoints surrounding the actor. 
Such human-centric free viewpoint video generation with bullet-time effects further evolves as a cutting-edge yet bottleneck technique with the rise of VR/AR over the last decade.
How to enable flexible and convenient bullet-time design for human free-viewpoint rendering with fully natural and controllable viewing experiences remains unsolved and has recently attracted substantive attention of both the multi-media and computer graphics communities.

To produce such bullet-time effect for dynamic human activities, early solutions~\cite{gortler1996lumigraph,carranza2003free,zitnick2004high} require dense and precisely arranged cameras which are aligned on a track forming a complex curve through space and restricted to pre-designed camera trajectory without the ability of convenient and flexible trajectory design.
The model-based solutions~\cite{Mustafa_2016_CVPR,collet2015high} rely on multi-view dome-based setup for high-fidelity reconstruction and texture rendering in novel views to support flexible bullet-time design.
However, the model reconstruction stages require two to four orders of magnitude more time than is available for daily interactive applications.
The recent volumetric fusion methods~\cite{UnstructureLan,robustfusion,tao2021function4d} have enabled much faster human reconstruction by leveraging the RGBD sensors and modern GPUs. 
But they are still restricted by the limited mesh resolution and fail to provide photo-realistic bullet-time effects.
The recent neural rendering techniques~\cite{Wu_2020_CVPR,NeuralVolumes,mildenhall2020nerf,tretschk2020nonrigid,NeuralHumanFVV2021CVPR,pumarola2020d} bring huge potential for neural human free-viewpoint rendering from multiple RGB input.
However, these solutions rely on per-scene training or are hard to achieve real-time performance due to the heavy network.
Only recently, some approaches~\cite{wang2021ibrnet,SRF} enhance the neural radiance field (NeRF)~\cite{mildenhall2020nerf} to break the per-scene training constraint, while other approaches~\cite{rebain2020derf,neff2021donerf,yu2021plenoctrees,garbin2021fastnerf,reiser2021kilonerf} further enable real-time NeRF rendering for static scenes. 
However, researchers did not explore these solutions to strengthen the bullet-time design for human activities, especially for the flexible, interactive and real-time setting without specific per-scene training.
Moreover, existing approaches~\cite{mildenhall2020nerf,wang2021ibrnet,SRF} refine the NeRF represenration using all the input frames, leading to inefficient training overload for dynamic scenes without exploiting the temporal redundancy, especially for our trajectory-aware bullet-time design task.

In this paper, we attack these challenges and present \textit{iButter} -- a neural \textbf{I}nteractive \textbf{BU}lle\textbf{T}-\textbf{T}ime G\textbf{E}ne\textbf{R}ator for human free-viewpoint rendering from multiple RGB streams.
As shown in Fig.~\ref{fig:teaser}, our approach enables real-time immersive rendering of novel human activities based on neural radiance field, which is critical for artists to design arbitrarily bullet-time trajectories interactively, flexibly and conveniently.
Once the trajectory designed, our approach further provide an efficient trajectory-aware refinement scheme within about 20 minutes to utilize the spatial and temporal redundancy for photo-realistic bullet-time rendering.

Generating such a human free-viewpoint video by combining bullet-time design with neural radiance field rendering in a real-time, general and dynamic manner is non-trivial.
More specifically, we adopt doom-like dense multiple RGB streams as input and utilize the neural radiance field~\cite{mildenhall2020nerf} as the underlying scene representation.
During the real-time preview stage for flexible bullet-time trajectory design, inspired by the general NeRF approach~\cite{wang2021ibrnet}, we adopt the network architecture with pixel-aligned feature extractor, implicit scene representation and radiance-based volumetric renderer so as to get rid of the tedious per-scene training, with the aid of a hybrid training set using both our doom system and the Twindom dataset.
For further real-time inference in our bullet-time design task, we utilize the inherent global information from our multi-view setting via a efficient silhouette-based sampling strategy. 
We also adopt a light-weight network design for each architecture components, which serves as a good compromising settlement between fast run-time performance and realistic preview quality.
With the interactively designed trajectory, we further introduce a novel efficient trajectory-aware refinement scheme to provide high-quality and photo-realistic bullet-time viewing experience along the trajectory.
To this end, we encode the spatial consistency by re-rendering the target image into an adjacent view through our refined network in a self-supervised manner.
We also encode the temporal consistency between two successive frames using the correspondences found by non-rigid deformation on the geometry proxies of the preview stage. 
Besides, we introduce a semantic feature aggregation scheme to enhance the rendering details for our refinement stage. 

\begin{figure*}[t]
\centering
	\centerline{\includegraphics[width=\textwidth]{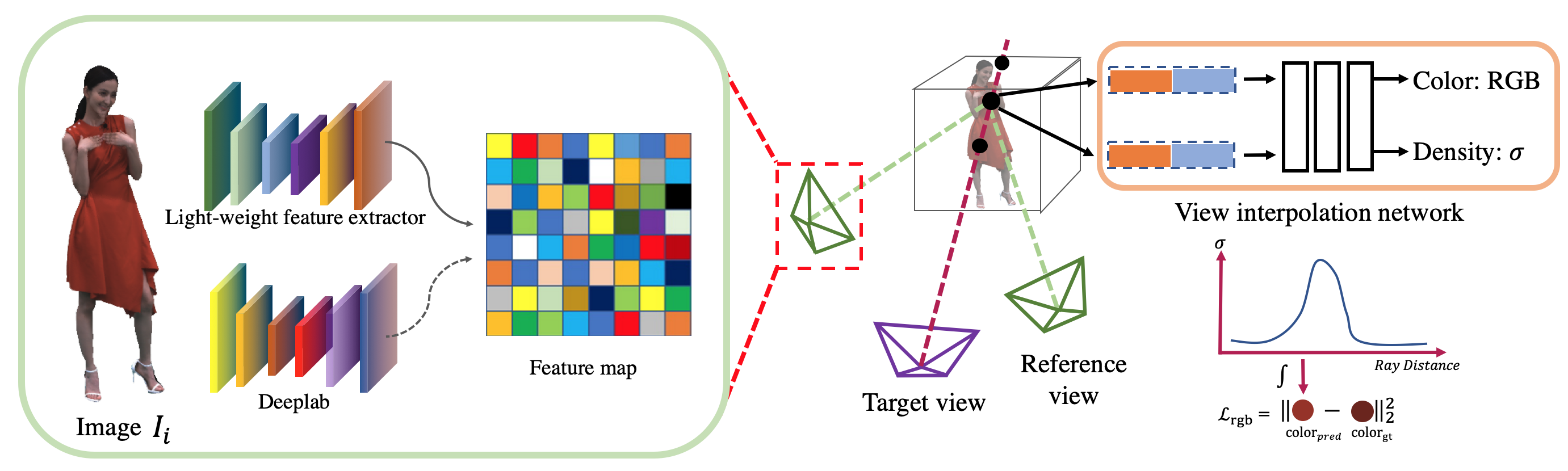}}
	\vspace{-10pt}
\caption{Illustration of our network architecture. Figure on the left demonstrates the feature extraction of our iButter. The Deeplab features (dot curve) only be used in high-quality bullet-time rendering stage. Features of sample points are fetched from reference views' feature maps.}

\label{fig:network}
\vspace{-4mm}
\end{figure*}

To summarize, our main contributions include:
\begin{itemize}
    \item We present a neural bullet-time generation scheme for photo-realistic human free-viewpoint rendering from dense RGB streams, which enables flexible, convenient and interactive design for human bullet-time visual effects unseen before.
    
    \item We propose a bullet-time preview approach which extends the NeRF rendering to a real-time, general and dynamic setting, with the aid of hybrid training set, light-weight network design and an efficient silhouette-based sampling strategy.
    
    \item We introduce a trajectory-aware refinement scheme in 20 minutes which jointly encodes the spatial, temporal consistency and semantic cues for photo-realistic bullet-time viewing experience along the designed trajectory.

\end{itemize}

\section{Related Work}

\myparagraph{Human Modeling.}
Modeling a human is a critical step in generating high quality free-viewpoint rendering.  Some works \cite{580394,collet2015high,Mustafa_2016_CVPR}  have reconstructed high-quality dynamic geometric models and texture maps by building complex camera systems and controllable lighting systems to overcome the occlusion challenges. The time issue of model reconstruction is preventing these methods from daily interactive applications. With the utility of the depth sensor and the powerful computing power of GPU, some works \cite{Newcombe2015, robustfusion,yu2018doublefusion, guo2017real, xu2019flyfusion} combine the volumetric fusion and non-rigid tracking to reconstruct free-form human body in real-time by using only a single RGBD sensor. However, the monocular RGBD methods are limited by the self-occlusion and the capture quality of the depth sensor and cannot reconstruct high-quality models and texture with details. Recently, some work\cite{kanazawa2018end, kolotouros2019convolutional, omran2018neural} has been used to apply deep learning methods to recover the geometry and texture of the human body from a single image.  Some work \cite{zheng2019deephuman, gabeur2019moulding, kanazawa2018end, kolotouros2019convolutional, omran2018neural, saito2019pifu,saito2020pifuhd} formulate the human body reconstruction problem into the regression problem of the shape and pose parameters of the parametric models. And recent studies combine deep learning and different 3D representations including depth maps \cite{gabeur2019moulding}, voxel grids \cite{zheng2019deephuman} and implicit functions\cite{saito2019pifu,saito2020pifuhd} to recover human model from single image. But these methods always fail to reconstruct  high-quality geometry and fine-detailed texture in order to achieve photo-realistic rendering in novel views.
Comparably,
our approach enables photo-realistic human free-viewpoint
video generation using only rgb inputs with mask information to generate SFS prior.

\myparagraph{Neural Rendering.}
The recent progress of neural rendering \cite{DeepBlending} shows great capacity to achieve photorealistic novel views synthesis based on various data representation, such as pointclouds \cite{Wu_2020_CVPR}, voxels\cite{Liu2018Neural} and  implicit function \cite{mildenhall2020nerf,NeuralHumanFVV2021CVPR,saito2019pifu,luo2021convolutional}.
For dynamic scenes neural rendering, NHR \cite{Wu_2020_CVPR} using dynamic point clouds to encode sequence feature.  
Neural Volumes \cite{NeuralVolumes} encodes multi-view images into a latent code and decodes to a 3D volume representation followed by volume rendering.
Recent work \cite{pumarola2020d, park2020deformable} use radiance field to render dynamic scenes. They regress each time step into a canonical frame. However, these methods need per-scene training is hard to provide an instant preview for an unseen scene. Other methods, \cite{Yu20arxiv_pixelNeRF,chen2021mvsnerf,SRF,wang2021ibrnet} try to extend the ability of neural radiance field to arbitrary new scenes. However, they cannot achieve real-time rendering to provide users with interactive bullet-time trajectory design and may have worse performance on the unseen scene than per-scene training models. While other methods \cite{neff2021donerf,garbin2021fastnerf,reiser2021kilonerf,yu2021plenoctrees, hedman2021snerg} try to render static scene in real-time but require per-scene training. In contrast, our method can generalize to new, dynamic, real-world scenes for interactive preview without long time training and a high-quality refinement in a short time. 

\myparagraph{Free Viewpoint Rendering.}
Traditional image-based rendering generate novel view texture by using the input images blending \cite{levoy1996light,carranza2003free,chen1993view,matusik2000image,zitnick2004high}. They calculate blending weights by considering camera view angles and spatial distance. Other methods \cite{buehler2001unstructured} use explicit geometry prior to guiding novel view rendering. However, these methods are limited by the accuracy of geometry prior. Another direction is using the multiplane images \cite{mildenhall2019llff, broxton2020immersive,choi2019extreme,srinivasan2019pushing} to represent the scene. While these methods cannot provide a wide range of free viewing rendering effects. Our approach can generate arbitrary free view rendering results from the user-designed bullet-time trajectory.

\vspace{-3mm}
\section{Overview}

We propose a \textbf{I}nteractive \textbf{BU}lle\textbf{T}-\textbf{T}ime G\textbf{E}ne\textbf{R}ator, or iButter, which allows users to interactively making photo-realistic human bullet-time effects only from multi-view RGB image sequences without heavy 3D reconstruction. 
The iButter's pipeline can be divided into two stages. The first stage is real-time dynamic 3D human previewing, and the second one is high-quality bullet-time rendering, which are described as follows.

\vspace{1mm}
\myparagraph{Real-time Dynamic 3D Human Previews.}
One key feature of our iButter is providing real-time previews for interactive virtual camera trajectory selection. 
And the selected trajectory will be applied to render high-quality bullet-time effects.
Our iButter adopts recent neural radiance fields work~\cite{wang2021ibrnet} to our real-time photo-realistic rendering task and achieves generic view interpolation directly based on RGB images without per-scene training and time-consuming 3D reconstructions.  
In section \ref{sec:network}, we first introduce the network architecture of iButter for generic view synthesis from multi-view video streams. 
Meanwhile, a silhouette-based sampling strategy leverages inherent global information from human silhouettes and is utilized to significantly reduce computational complexity and accelerate neural training and rendering speed (section~\ref{sec:preview}). 
Accordingly, users can design bullet-time trajectories interactively and get intuitive feedback from our renderer.

\vspace{1mm}
\myparagraph{High-quality Bullet-time Rendering.}
Moreover, iButter can produce high-resolution photo-realistic rendering results with temporal and multi-view consistency on the given selected trajectory from users. 
To this end, we design a novel efficient trajectory-aware refinement scheme which is introduced in section~\ref{sec:refinement}. 
On the one hand, in the scheme, we enhance multi-view consistency by re-rendering a near reference camera with a rendered camera on the trajectory. 
On the other hand, we warp two adjacent rendered images on the trajectory according to model correspondences and force them to be the same for strengthening the temporal consistency of our iButter.
Also, we introduce a semantic feature aggregation scheme to enhance rendering quality and network generalization of human-related scenes. 
After these self-supervised refinements, iButter is able to render high-quality images on the selected trajectory and preservers multi-view and temporal consistency.

\section{Image-based Bullet-time Generator} \label{Sec:IBUTTER}

Here, we introduce our neural image-based bullet-time generator (iButter).
Firstly, we show the neural rendering network's architecture which can achieve generic photo-realistic view interpolation. 
Based on this kind of network, we introduce a silhouette-based sampling strategy for fast training, rendering, and previewing. 
We also propose a semantic feature aggregation scheme to improve the rendering results of human-related scenes.
Then user-selected trajectory will guide the trajectory-aware network refinement and help to enhance temporal and spatial consistency in a self-supervised manner.

\subsection{Network Architecture} \label{sec:network}

The IBRNet~\cite{wang2021ibrnet} synthesizes the novel target view by interpolating nearby source views instead of encoding an entire scene into a single network, which enables realistic rendering without per-scene training. 
We adopt this architecture to our bullet-time task with dynamic human and real-time free-view rendering. 
iButter evolves from this kind of architecture. The whole architecture can be divided into three parts: feature extraction, view interpolating, and volume rendering.
Specifically, in the beginning, iButter selects $K$ nearest reference cameras according to a given target camera's parameters $\mathbf{P}_v$. 
In the first part, we extract features from input images. Let $\mathbf{I}_i$ denotes the color image of $i$-th source view at one frame. We use a feature extractor $\Phi_f(\cdot)$ to extract feature maps from RGB images $\{\mathbf{I}_i\}$ . 
So we have a set of cameras and feature maps $\mathcal{S}=\{(\mathbf{P}_i, \Phi_f(\mathbf{I}_i))\}_{i=1}^K$, where $\mathbf{P}_i$ is the camera's parameters of $i$-th source view. 
In the view interpolating part, iButter fetches features from $\mathcal{S}$ for each sample points along camera rays and predicts their density $\sigma$ and color $\mathbf{c}$ via our view interpolation network in a hierarchical manner. 
Different from IBRNet, we replace the ray transformer with simple MLPs in the view interpolation network. 
%
Note that the output of the view interpolating part is colors and densities of sample points along rays. 
Finally, to render the color of a ray $\mathbf{r}$ through the scene, we accumulate colors along the ray modulated by densities in the volume rendering part. 
The volume rendering part is formulated as :
\begin{equation}
\label{equ:volumerendering}
\begin{split}
\tilde{C}(\mathbf{r}) &=\sum_{k=1}^{n_s} T_k(1-exp(-\sigma_k\delta_k))\mathbf{c}_k, \\
T_k &= exp(-\sum_{j=1}^{k-1}\sigma_j\delta_j),
\end{split}
\vspace{-20pt}
\end{equation}
where $n_s$ is the number of sample points along the ray $\mathbf{r}$. $\sigma_k$, $\mathbf{c}_k$, $\delta_k$ denotes the density, color of the samples and the interval between adjacent samples respectively.

The architecture of our iButter is illustrated in Fig.~\ref{fig:network}. Please refer to~\cite{wang2021ibrnet} for more details. 

%

\begin{figure}[t]
\centering
	\centerline{\includegraphics[width=.5\textwidth]{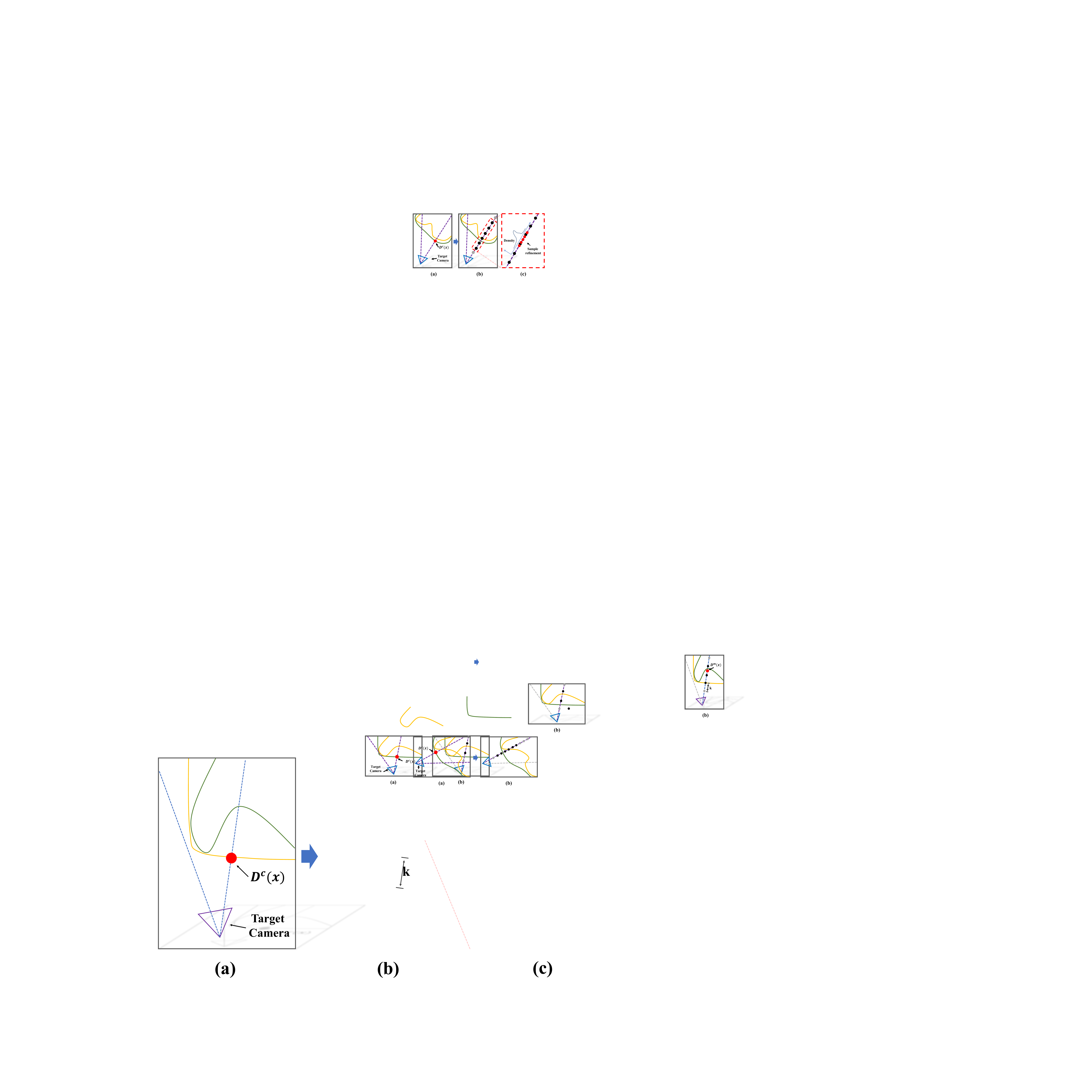}}
		\vspace{-10pt}
\caption{Demonstration of our silhouette-based sampling strategy. The orange curve represents the real geometry surface; The green curve is the surface generated by SfS. (a) iButter calculates depth values for camera rays on coarse geometry $\mathbf{G}$; (b) Sampled points near the depth value in "coarse" stage; (c) Sampled points according to density distribution in "fine" stage.}
\label{fig:sfs}
\vspace{-4mm}
\end{figure}

\begin{figure*}[t]
\centering
	\centerline{\includegraphics[width=\textwidth]{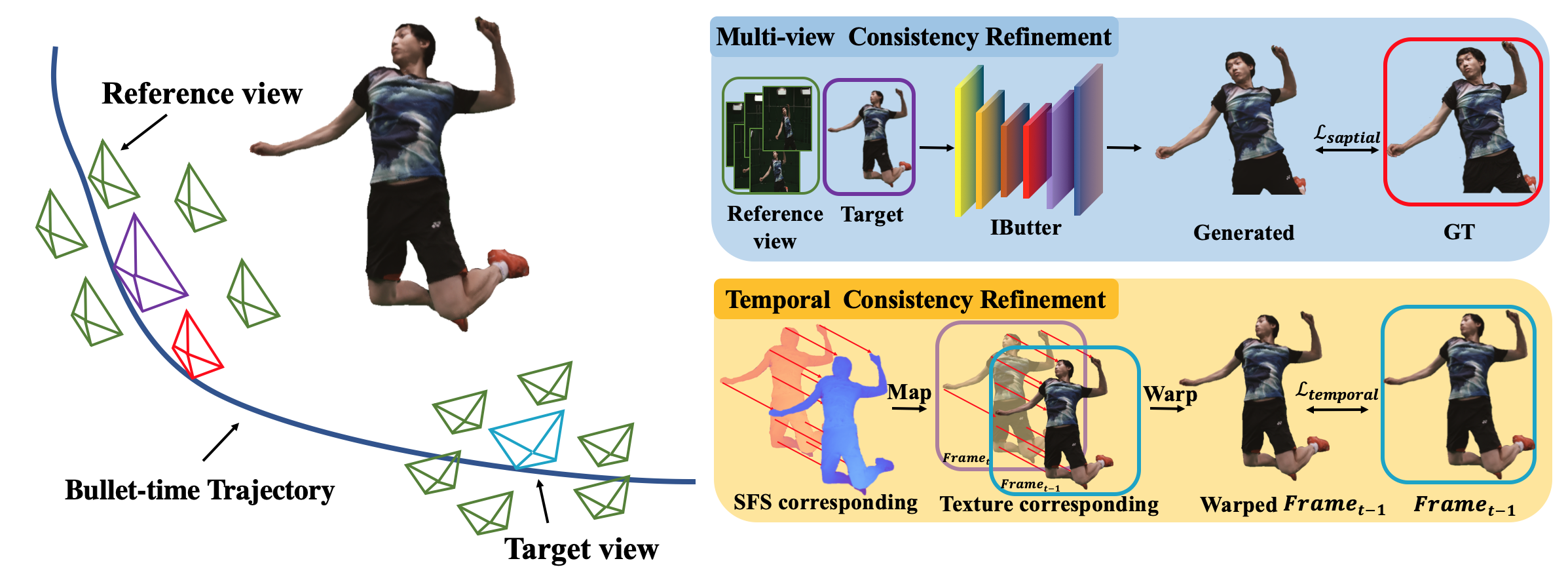}}
		\vspace{-10pt}
\caption{Illustration of our trajectory-aware refinement pipeline. Both spatial consistency loss and temporal consistency loss are applied to improve rendering quality.
}
\label{fig:traj1}
\vspace{-4mm}
\end{figure*}

\subsection{Real-time Previews and Trajectory Selection}\label{sec:preview}

According to Equation~\ref{equ:volumerendering}, colors of sample points near human geometry surface contribute primarily to the ray color $\tilde{C}(\mathbf{r})$ in an ideal radiance field. 
Weights for sample points that are away from the surface will be close to zeros.
So based on this observation, instead of sampling all points between near-far planes along with rays, we propose a sampling strategy based on multi-view silhouettes to sample around human geometry surface efficiently. 
This strategy can significantly accelerate both training and rendering speed by reducing useless sample points.

Specifically, for each frame with multi-view RGB image data, we generate a coarse 3D human mesh $\mathbf{G}$. 
We deploy a fast Shape-from-Silhouette (SfS) method~\cite{yous2007gpu} to reconstruct geometry from multi-view human silhouettes in real-time. 
Having green screens in our multi-view capture system, we can fast extract human silhouettes from RGB images via chroma key segmentation and background subtraction.

Before point sampling along a ray $\mathbf{r}$, we first calculate the depth value from reconstructed $\mathbf{G}$. 
Once the ray $\mathbf{r}$ hits human geometry, we evenly sample points near the depth value within a small range in the coarse sampling stage and then follow the hierarchical sampling scheme in~\cite{mildenhall2020nerf}. 
Fig.~\ref{fig:sfs} illustrates our silhouette-based sampling strategy. 
Besides, in the preview stage, we deploy a UNet-like light-weight network as the feature extractor $\Phi_f$, which has a downsampling scale of 4. Please refer to the supplementary material for more network details.
We formulate iButter $\Phi(\cdot)$ as:
\begin{equation}
\begin{split}
\mathbf{\tilde{I}} = \Phi(\mathcal{S}, \mathbf{P}_v, \mathbf{G}),
\end{split}
\vspace{-20pt}
\end{equation}
where $\mathbf{\tilde{I}}$ is the rendered image in the target view.

All these efficient procedures enable a real-time rendering from multi-view live streams. 
With our iButter, users can select trajectory presets and adjust key frames through an user interface (UI) in an interactive way with instant feedbacks, which is artist friendly.

\subsection{High-quality Bullet-time Rendering} \label{sec:refinement}

Although the real-time version of iButter has achieved rather realistic rendering result, the synthesis quality is good but not enough for production. 
To this end, we propose feature aggregation and trajectory-aware refinement to boost rendering quality into a product level. 
Similarly, we also bring our silhouette-based sampling strategy introduced in \S~\ref{sec:preview} into our high-quality bullet-time rendering so as to accelerate both training and rendering speed.

\vspace{3mm}\noindent{\bf Semantic Feature Aggregation.}
As mentioned before, iButter produces a discrete radiance field according to fetched features of sample points from multi-view image feature maps. 
Ideally, more discriminative features will facilitate networks to approximate human geometry and appearance more correctly, produce a more accurate discrete radiance field, and better rendering results. 
Here, we aggregate both local features from our light-weight feature extractor and high-level semantic features from deeplab~\cite{chen2017deeplab} segmentation pre-trained backbone to form a new discriminative feature extractor $\Phi_f$, as illustrated in Fig.~\ref{fig:network}. 
Although increasing the computational complexity, the new feature extractor boosts the rendering quality and is one step closer to product-level results as discussed in section~\ref{sec:eva} .

\vspace{3mm}\noindent{\bf Trajectory-aware Refinement.}
Given selected bullet-time trajectory, we set out to refine iButter for product-level rendering results on it. 
Multi-view consistency and temporal consistency are two main challenges for such neural renderers like ours because iButter predicts radiance fields according to view image features instead of maintaining a constant representation.  
Without these consistencies, iButter will produce flickering and ghosting artifacts on result images. 
We attack these two challenges by introducing a training scheme to alleviate such artifacts based on given trajectory. 
Figure~\ref{fig:traj1} demonstrates the pipeline of our trajectory-aware refinement.  

The selected bullet-time trajectory defines $K$ related reference cameras in each frame. 
So we can narrow down the training dataset by discarding other unrelated cameras. 
For enhancing multi-view consistency, we first render the target view image $\mathbf{\tilde{I}}_v^t= \Phi(\mathcal{S}^t, \mathbf{P}_v^t, \mathbf{G}^t)$ in the frame $t$. 
Next, we utilize $\mathbf{\tilde{I}}_v^t$ to a re-render reference view with other $K-1$ reference cameras, and supervise the network training with the ground truth. 
We formulate the multi-view consistency loss $\mathcal{L}_{spatial}$ as:
\begin{equation}
\begin{split}
\mathcal{L}_{spatial} = \frac{1}{2}\sum_{t=1}^T\sum_{i=1}^K \| \mathbf{M}_v^{t}\cdot(\mathbf{I}_i^t - \Phi(\mathcal{S}^t\cup s_v^t \setminus s_i^t, \mathbf{P}_i^t, \mathbf{G}^t)) \|_2^2, \\
\end{split}
\end{equation}
where $s_v^t = (\mathbf{P}_v^t, \Phi_f(\mathbf{\tilde{I}}_v^t))$ is the generated camera-feature pair for target view camera on the trajectory, and $T$ is the length of bullet-time trajectory; $\mathbf{I}_i^t$ is the ground truth image of reference view $i$. $\mathbf{M}_v^{t}$ is the human foreground mask generated by the coarse geometry at frame $t-1$ for avoiding learning background pixels.

\begin{figure*}[t]
\centering
	\centerline{\includegraphics[width=.85\textwidth]{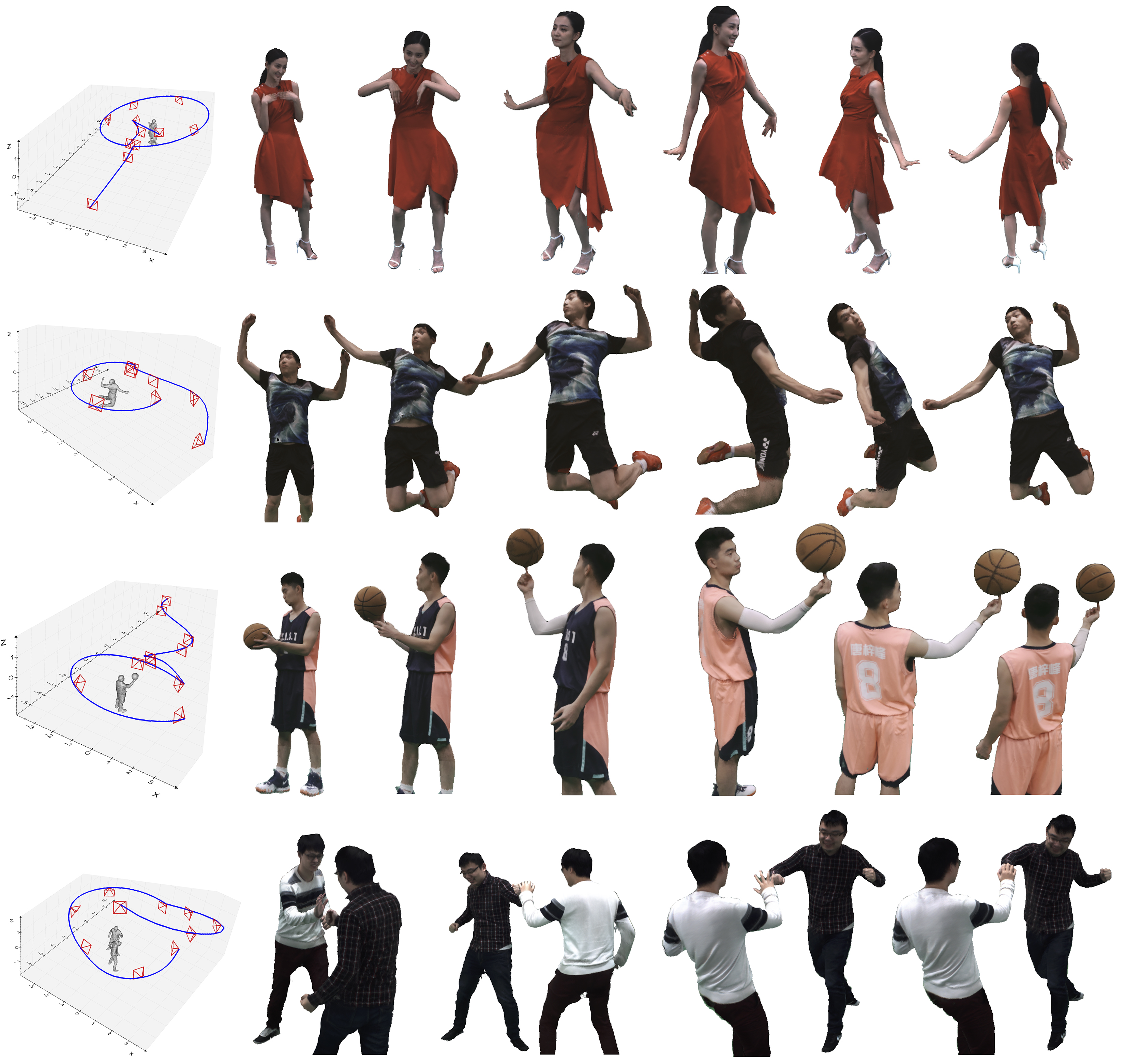}}
	\vspace{-10pt}
\caption{Several examples that demonstrate the quality and fidelity of the render results (right) from the trajectory user designed (left) from our system on the data we captured, including human portrait, human with objects and multi humans.}

\label{result_gallery}
\vspace{-4mm}
\end{figure*}

Since having coarse human geometry for each frame, we find correspondences between adjacent 3D models on timeline by a light-weight model tracking algorithm~\cite{xu2019unstructuredfusion}. 
These model correspondences establish connections between these two adjacent frames and are used for image warping.
Let $f_{t, t-1}(\cdot)$ represents the 2D flow which warps image from frame $t$ to frame $t-1$ and projected from model correspondences. 
We formulate the temporal consistency loss $\mathcal{L}_{temporal}$ as:
\begin{equation}
\begin{split}
\mathcal{L}_{temporal} = \frac{1}{2}\sum_{t=1}^T\| \mathbf{M}_v^{t-1}\cdot(\mathbf{\tilde{I}}_v^{t-1} - f_{t, t-1}(\mathbf{\tilde{I}}_v^t))\|_2^2, \\
\end{split}
\end{equation}
where $\mathbf{\tilde{I}}_v^{t-1}$ and $\mathbf{\tilde{I}}_v^t$ are target view images generated by iButter for adjacent frames.

\section{Implementation Details} \label{sec:imp}

Our multi-view data are captured from a dome system with up to 80 cameras arranged on a cylinder. All cameras are synchronized and capture at 25 frames per second. 
For training data, we collect 20 sets of datasets where the performers are in different clothing and perform different actions.
Moreover, we use Twindom dataset~\cite{twindom} to generate multi-view synthetic datasets with single frame to extend training set in diversity. 
We first pretrain iButter both the feature extraction network and the view interpolation network end-to-end on training set using a single photometric loss:
\begin{equation}
\begin{split}
\mathcal{L}_{rgb} = \frac{1}{2}\sum_{t=1}^T\sum_{i=1}^N \| \mathbf{M}_v^{t}\cdot(\mathbf{I}_i^t -\mathbf{\tilde{I}}_i^t) \|_2^2, \\
\end{split}
\end{equation}
where $N$ is the number camera in the frame. 

During trajectory-aware refinement, we combine all losses together and formulate total loss $\mathcal{L}$ as:
\begin{equation}
\begin{split}
\mathcal{L} = \lambda_1\mathcal{L}_{rgb} + \lambda_2\mathcal{L}_{spatial} + \lambda_3\mathcal{L}_{temporal},
\end{split}
\end{equation}
where $\lambda_1=0.5$ and $\lambda_2 = \lambda_3 = 0.25$ in this paper. 
Note that we do not update the parameters of deeplab backbone during refinement. 

We use Adam optimizer to train our networks with a base learning rates of $10^{-3}$ and $5\times10^{-4}$ for feature extraction network and the view interpolation network respectively.

\begin{figure*}[t]
	\centering
	\centerline{\includegraphics[width=\textwidth]{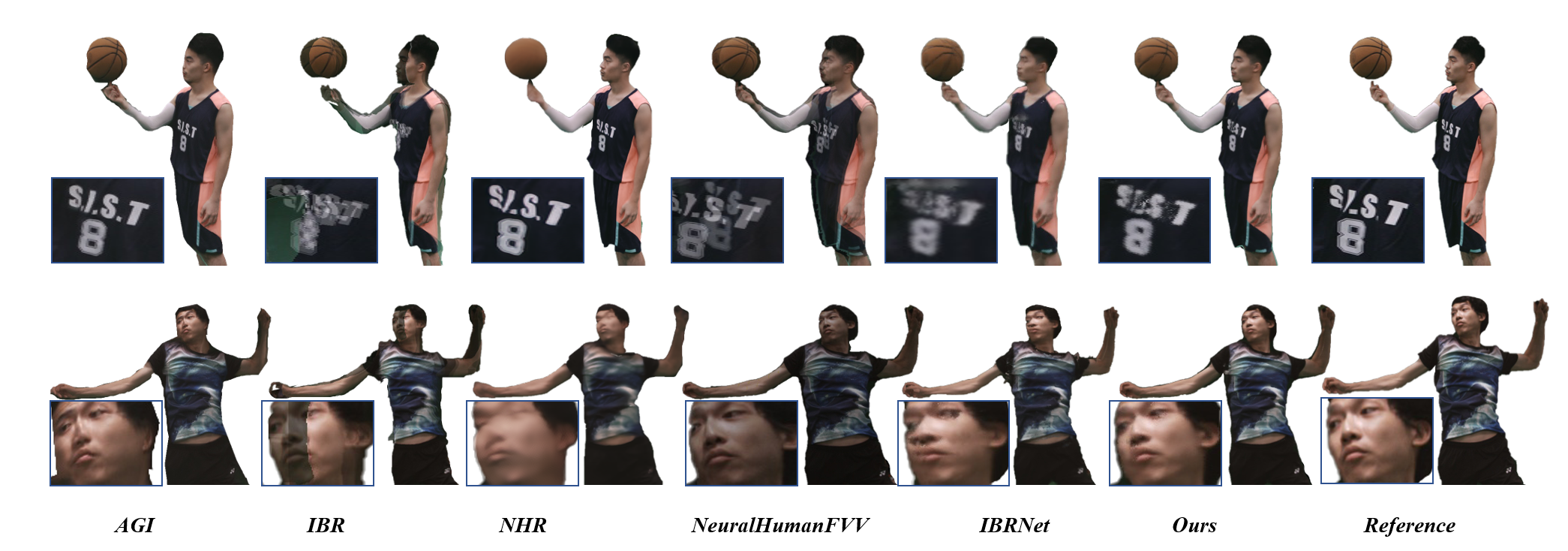}}
	\vspace{-10pt}
	\caption{Comparison of our methods with AGI \cite{Agi}, IBR \cite{10.1145/383259.383309}, NHR \cite{Wu_2020_CVPR}, 
		NeuralHumanFVV \cite{NeuralHumanFVV2021CVPR}, and IBRNet \cite{wang2021ibrnet}. Our neural portrait achieves much more realistic rendering results.}
	\label{Comparison}
	\vspace{-4mm}
\end{figure*}
\section{Experimental Results}
In this section, we evaluate the result of our iButter system on a variety of scenarios followed by the comparison with other methods, both qualitatively and quantitatively. 
We pre-train our model on a Nvidia GeForce RTX3090 GPU for 3 days with 400 rays per batch and it takes only 20 minutes for our trajectory-aware refinement using the same GPU format.
We sample 10 points in the coarse stage during the preview bullet-time design stage and sample 64 points in the coarse stage and additional 64 points in the fine stage during refinement. 
We render images with the output resolution of $512\times 384$ in our interactive preview using about 0.15 seconds and $1024\times 768$ in our final refinement result using about 30 seconds.
Fig.~\ref{result_gallery} provides some representative results to demonstrate that our approach generates photo-realistic human free-viewpoint rendering from dense RGB streams, which enables flexible and interactive design for human bullet-time visual effects. Please refer to the supplementary video for more video results.

\begin{table}[t]
	\centering
	\begin{tabular}{@{}c|cccccc@{}}
		\toprule
		Method & AGI          & IBR         & NHR   & FVV & IBRNet & Ours       \\ \midrule
		PSNR $\uparrow$        & 22.99          & 28.66         & 31.56    & 23.29 & 29.41 & \textbf{32.27}    \\
		SSIM $\uparrow$ &.9489 & .9661 & .9687 &.9203 &.9572 & \textbf{.9716}\\
		LPIPS $\downarrow$ &.0716 & .0749 &\textbf{.0644} &.0863 &.0846&.0659\\ \bottomrule
	\end{tabular}
	\caption{Quantitative comparsion against various methods.}
	\label{comp_tab}
	\vspace{-10pt}

\end{table}

\subsection{Comparison}
For thorough comparison, we perform quantitative and qualitative comparisons against \textbf{AGI} \cite{Agi} using the Agisoft software, the image-based method \textbf{IBR}~\cite{10.1145/383259.383309}, the point-based \textbf{NHR}~\cite{Wu_2020_CVPR}, the NeRF-basded \textbf{IBRNet}~\cite{wang2021ibrnet} and the hybrid texturing-based \textbf{NeuralHumanFVV}~\cite{NeuralHumanFVV2021CVPR}.
%
%
%
Note that NeuralHumanFVV and IBRNet are trained using the same training dataset as ours and IBR adopts the same SFS geometry proxy as ours for fair comparison.
As shown in Fig.~\ref{Comparison}, the non-learning approaches AGI and IBR suffer from severe texturing artifacts due to inaccurate geometry reconstruction.
Besides, NeuralHumanFVV and IBRNet suffer from mis-blending or blur artifacts, especially for the fast motion regions.
The NHR achieves promising results in those slow-motion scenarios loses fidelity in fast motions due to the ambiguity of point feature learning.
Moreover, it takes 7 days to train NHR on a specific sequence to achieve reasonable texturing results.
In contrast, our approach achieves photo-realistic free-viewpoint rendering of human activities especially for those fast motion regions, which only needs less than 20 minutes for refinement.
Our approach also enables human bullet-time visual effects design in a flexible and interactive manner, based on our real-time, dynamic NeRF-based rendering without per-scene training. 
For quantitative comparison, we adopt Peak Signal-to-Noise Ratio (\textbf{PSNR}), Structural Similarity (\textbf{SSIM})~\cite{SSIM} and Learned Perceptual Image Patch Similarity (\textbf{LPIPS})~\cite{LPIPS} as metrics on the whole testing dataset by comparing the rendering results with source view inputs.
As shown in Tab.~\ref{comp_tab}, our approach outperforms other methods in terms of PSNR and SSIM and achieves comparable performance against NHR in terms of LPIPS.
These comparisons above illustrate the effectiveness of our approach for flexible bullet-time design and photo-realistic free-viewpoint rendering. More comparison details can be found in supplementary materials.

\subsection{Evaluation} \label{sec:eva}
\begin{figure}[t]
	\centering
	\centerline{\includegraphics[width=.5\textwidth]{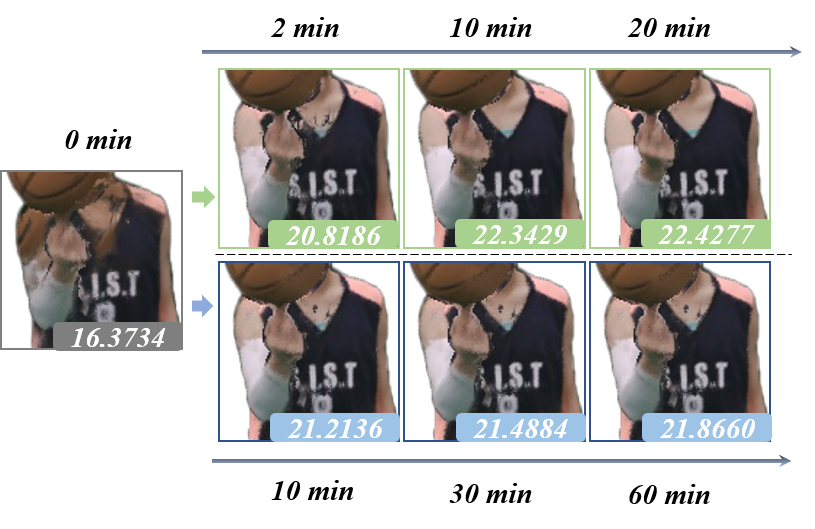}}
	\vspace{-10pt}
	\caption{Evaluation of Trajectory-based Refinement. Our trajectory-based scheme (top) achieves more efficient and effective refinement than the brute force one (bottom).Note that our results with in 20 minutes are better than the results of brute force scheme in 60 minutes. PSNR of each cropped image is attached in the button of sub-figure.}
	\label{eva3}
	\vspace{-4mm}
\end{figure}

\begin{figure}[t]
	\centering
	\centerline{\includegraphics[width=.5\textwidth]{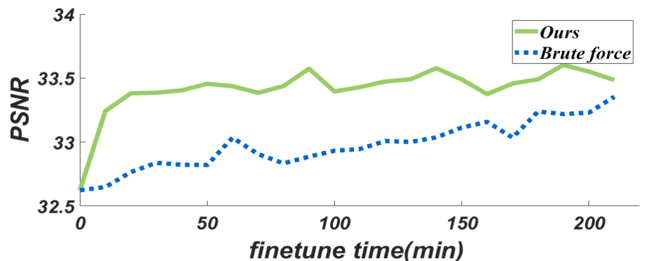}}
	\vspace{-10pt}
	\caption{Quantitative evaluation of the refinement stage in terms of refining time.}
	\label{eva32}
	\vspace{-4mm}
\end{figure}

\vspace{1mm}\noindent{\bf Trajectory-based Refinement.}
Here we compare our trajectory-based refinement scheme with the brute force scheme using all the input frames (denoted as \textbf{Brute force}). 
As illustrated in Fig.~\ref{eva3}, our trajectory-aware scheme achieve more efficient refinement within 20 minutes by jointly utilizing the spatial and temporal redundancy for the bullet-time design task.
Fig.~\ref{eva32} further provides the final PSNR with various refining time for two schemes.
While brute force scheme suffers from inefficient refinement as ours even with longer refining time, our scheme achieve more efficient and temporal consistent rendering results.

\vspace{1mm}\noindent{\bf Number of sample points and feature extraction network.}
Here we evaluate the number of per-ray sample points and feature extraction network architecture in our interactive preview stage by rendering 200 novel views on a 24GB Nvidia GeForce RTX3090 at the resolution of $512\times 384$. 
As shown in Tab.~\ref{table:num_points} (see supplementary material
for more qualitative and quantitative results), more sample points and deeper feature extractor lead to better rendering quality and more rendering time.
Through these thorough evaluations, we choose to sample 10 points per-ray and 3 residual blocks in feature extraction in the interactive preview stage, which serves as a good compromising settlement between fast run-time performance and flexibly interactive ability for bullet-time design.

\vspace{1mm}\noindent{\bf Ablation study of refinement network.}
Here we evaluate the design of our refinement network. 
Let \textbf{w/o SFS}, \textbf{w/o SFS}, \textbf{w/o semantics}, \textbf{w/o multi-view}, \textbf{w/o temporal} denote our variations without using the SFS proxy, the semantics feature aggregation, the spatial multi-view supervision and the temporal supervision, respectively. 
 Tab.~\ref{eval1_t} show that our full refinement scheme consistently outperforms other variations in terms of all the metrics (see supplementary material
for qualitative comparison). 
These evaluations not only highlights the contribution of each algorithmic component but also illustrates that our refinement enables photo-realistic human free-viewpoint rendering for bullet-time design.

\subsection{Limitation and Discussion}
As a trial to provide an interactive bullet-time generator for unseen scenes, our iButter is subject to some limitations. 
First, our approach still relies on about 80 cameras to provide a 360-degree free-view rendering.
It’s promising to reduce the camera number using human priors and even enabling in-the-wild capture with more  data-driven scene modeling strategies. 
Secondly, there still exists some flicking artifacts near the boundary regions due to the challenging viewing directions and some complex self occlusion areas. This could be alleviated in the future by assigning weights of source views using the visibility information from SFS prior.
Currently, our iButter achieves interactive dynamic 3D human preview in around 0.15 seconds. 
Speeding up our rendering for more fluent interaction is the direction we need to explore. 
Furthermore, our approach relies on a consistent lighting assumption and it’s promising to handle complex lighting conditions for view-dependent rendering. 
It's also interesting to model the changes between the adjacent frames, so as to enable motion interpolation effects for our users.

\begin{table}[t]
	\centering
	
	\begin{tabular}{@{}c|cccc@{}}
		\toprule
		\# Sample points & PSNR   $\uparrow$        & SSIM    $\uparrow$        & LPIPS    $\downarrow$   & rendering time (s)    $\downarrow$      \\ \midrule
		5          & 34.09          & .9800          & .0328  &  .1293       \\
		10         & 34.61          & .9822          & .0263 & .1529      \\
		20         & 34.91          & .9833          & .0241 & .2429          \\
		40         & 34.90          & .9828          & .0248 & .3258          \\
		60         & 35.01 & .9832 & .0239 & .9423\\ \bottomrule
	\end{tabular}
	\caption{Quantitative evaluation on the number of the sample points in terms of various metrics.
	}
	\label{table:num_points}
	\vspace{-3mm}
	\vspace{-10pt}
\end{table}

\begin{table}[t]
\centering

\begin{tabular}{@{}c|ccc@{}}
\toprule
                  & PSNR    $\uparrow$        & SSIM    $\uparrow$        & LPIPS   $\downarrow$       \\ \midrule
Ours               & \textbf{32.28} & \textbf{.9716} & \textbf{.0659} \\
w/o SFS    & 26.62          & .9259          & .0934          \\
w/o semantic     & 28.38          & .9413          & .0821          \\
w/o multiview     & 27.42          & .9404          & .0898          \\
w/o temporal     & 27.82          & .9424          & .0895          \\
 \bottomrule
\end{tabular}
\caption{Quantitative ablation study of refinement network.}
\label{eval1_t}
\vspace{-3mm}
\vspace{-10pt}
\end{table}

\section{Conclusion}
We have presented iButter, a neural interactive bullet-time generator for photo-realistic human free-viewpoint rendering from dense RGB streams, which enables human bullet-time visual effects design in a flexible and interactive manner unseen before.
%
Our real-time preview and design stage enables real-time and dynamic NeRF rendering of novel human activities without per-scene training, with the aid of a hybrid training set, light-weight network design and an efficient silhouette-based sampling strategy.
Our efficient trajectory-aware refinement stage further enables photo-realistic bullet-time viewing experience of human activities by jointly utilizing the spatial, temporal consistency and semantic cues along the designed trajectory.
Extensive experimental results demonstrate the effectiveness of our approach for convenient interactive bullet-time design and photo-realistic human free-viewpoint video generation, which compares favorably to the state-of-the-art.
We believe that our approach renews the bullet-time design for human free-viewpoint videos generation with more flexible and interactive design ability.
It serves as a critical step for interactive novel view synthesis, with many potential applications for fancy visual effects in VR/AR, gaming, filming or entertainment.

\section{Acknowledgments}
This work was supported by NSFC programs (61976138, 61977047), the National Key Research and Development Program (2018YFB2100 
500), STCSM (2015F0203-000-06), SHMEC (2019-01-07-00-01-E00003) and Shanghai YangFan Program (21YF1429500).

\bibliographystyle{ACM-Reference-Format}
\bibliography{sample-base}


\clearpage

	\fancyhead{}
	\section{APPENDIX}
	\appendix

	\section{Feature Extraction Network Architecture}
	
	\begin{figure}[h]
		\centering
		\centerline{\includegraphics[width=.35\textwidth]{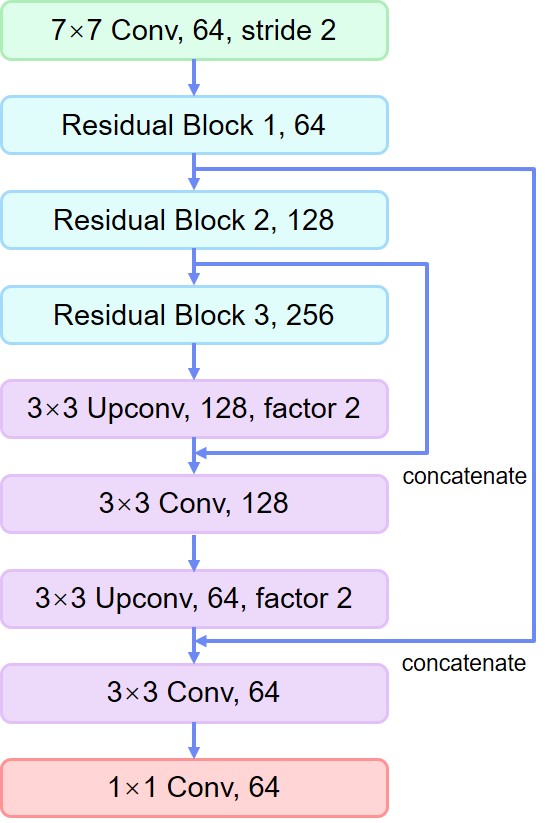}}
		\vspace{-10pt}
		\caption{Light-weight Feature Extraction Network Architecture.} 
		\label{Feature}
		\vspace{-2mm}
	\end{figure}
	
\begin{figure}[t]
		\centering
		\centerline{\includegraphics[width=.4\textwidth]{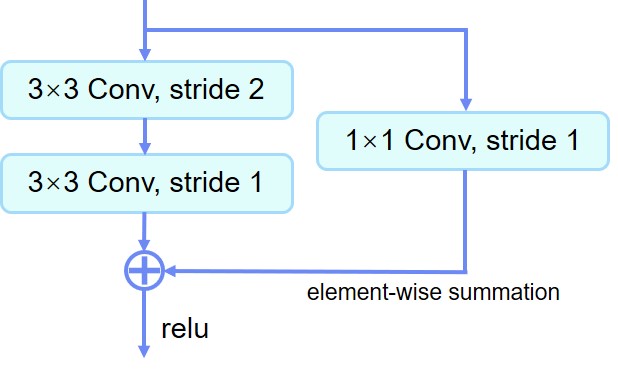}}
		\vspace{-10pt}
		\caption{Architecture for the Residue Block.} 
		\label{Residue}
		\vspace{-5pt}
	\end{figure}

	Fig. \ref{Feature} shows an overview of our light-weight feature extractor network architecture, which is modified from \cite{He_2016_CVPR} and \cite{wang2021ibrnet}. 
	Our network takes the source view as inputs and extracts their features using shared weights. “Conv" contains convolution, ReLU and Instance Normalization \cite{ulyanov2016instance} layers. 
	"Upconv" contains a bilinear upsampling with the factor and a "Conv" layer with stride 1. The detailed architecture of Residual Block is shown in Fig. \ref{Residue}. Compared to ResNet34 \cite{He_2016_CVPR}, we reduce convolution layers to speed up without losing performance in novel view synthesis.
	
	\section{Additional Results}
	\begin{figure*}[t]
		\centering
		\centerline{\includegraphics[width=\textwidth]{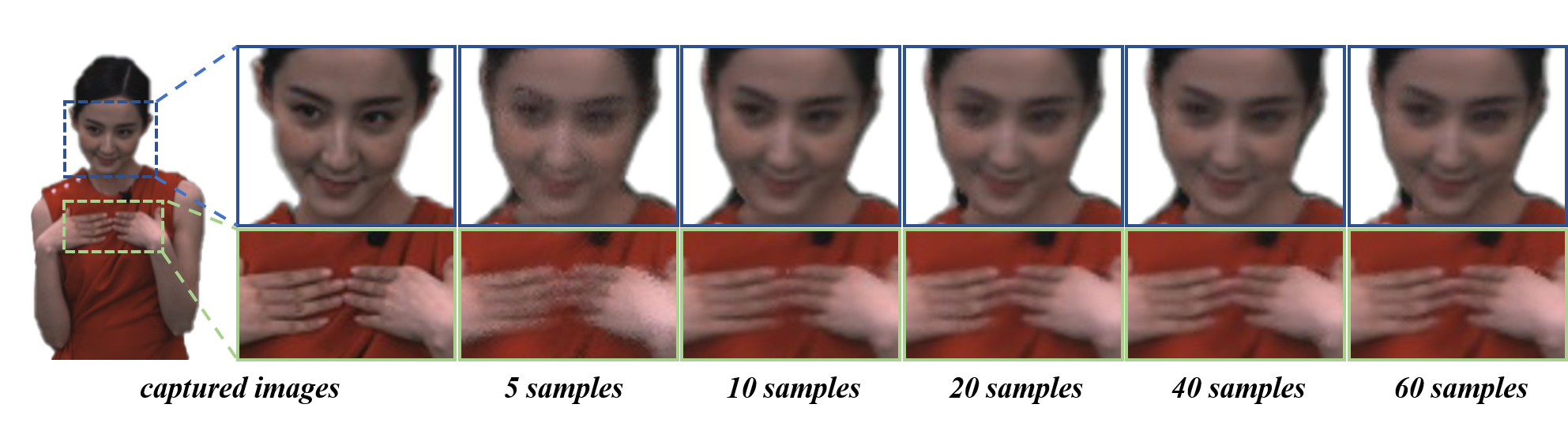}}
		\vspace{-10pt}
		\caption{Evaluation of number of sample points in our interactive renderer. Blur and artifacts will appear when only 5 points are sampled. When the number of sampling points exceeds 10, the performance of image does not increase much.} 
		\label{Number_Points}
		\vspace{-5pt}
	\end{figure*}
	
	\begin{figure*}[t]
		\centering
		\centerline{\includegraphics[width=1\textwidth]{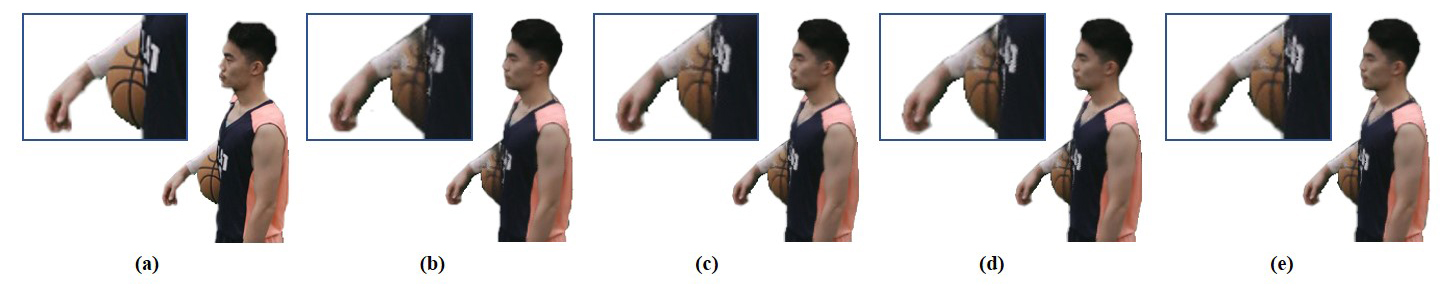}}
		\vspace{-10pt}
		\caption{Qualitative ablation study of feature extraction network. (a) ground truth; (b) using two residual blocks and one 'Upconv' layer ; (c) ours (three residual block); (d) using six residual blocks; (e) using nine residual blocks.}
		\label{supp}
		\vspace{-2mm}
		\vspace{-5pt}
	\end{figure*} 
	
	\begin{figure*}[t]
		\centering
		\centerline{\includegraphics[width=1\textwidth]{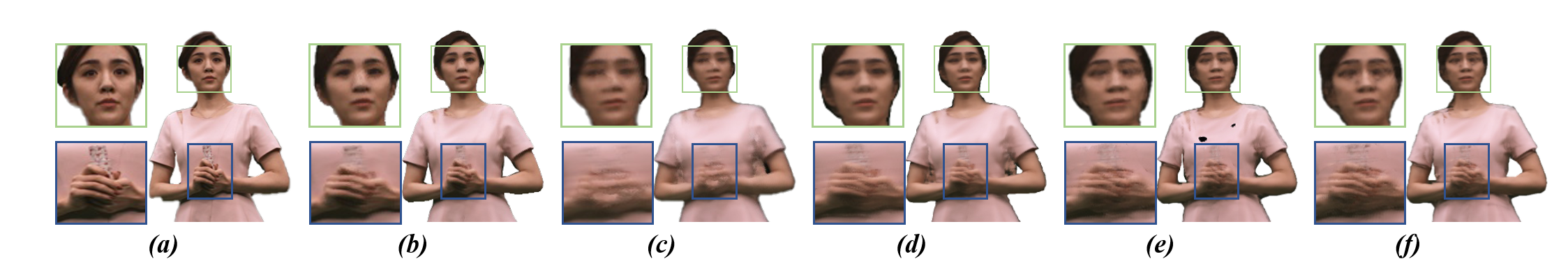}}
		\vspace{-10pt}
		\caption{Qualitative ablation study of refinement network. (a) ground truth; (b) ours; (c) w/o SFS prior; (d) w/o semantic feature aggregation; (e) w/o multi-view consistency refinement; (f) w/o temporal consistency refinement.}
		\label{eval1}
		\vspace{-5pt}
	\end{figure*}

	\subsection{Comparison}
	For the adopted implementations, in AGI we adopt the commercial Agisoft Metashape Professional software while in IBR we reproduce the traditional paper called Unstructured Lumigraph Rendering using the same SFS geometry proxy.
	We utilize the released official implementations for both NHR and IBRNet, and provide our data to the authors of NeuralHumanFVV for both training and testing.

	For the real-time performance of iButter, we render 200 512*384 images on an Nvidia GeForce RTX3090 GPU in average as shown in Tab.  \ref{comp_tab}. 
	Compared to the non-real-time methods (NeRF \cite{mildenhall2020nerf}, IBRNet \cite{wang2021ibrnet}, NHR \cite{Wu_2020_CVPR}), we can render at an interactive rate without per-scene training.
	Compared to those real-time methods (AGI \cite{Agi} built textured mesh, IBR \cite{10.1145/383259.383309}, NeuralHumanFVV \cite{NeuralHumanFVV2021CVPR}), we can provide photo-realistic and dynamic FVV preview without tedious per-scene training or reconstruction.

	\subsection{Evaluation}
	\vspace{1mm}\noindent{\bf Number of sample points and feature extraction network.}
	We provide qualitative evaluation of our models for using different number of sample points per ray in Fig. \ref{Number_Points}. We provide qualitative and quantitative evaluation of our feature extraction network in Fig. \ref{supp} and Tab. \ref{table:feature}.

	
	
	\begin{table}[t]
		\centering
		\begin{tabular}{@{}c|cccccccc@{}}
			\toprule
			Method & Ours &NeRF & IBRNet & NHR &AGI & IBR           & FVV        \\ \midrule
			time(s)   &.1529s & 26.78s & 21.33s & .3217s & .0071s    & .0253s & .0462s    \\
			\bottomrule
		\end{tabular}
		\caption{Quantitative rendering time comparsion against various methods.}
		\label{comp_tab}
		\vspace{-10pt}
	\end{table}
	
	\begin{table}[t]
		
		\centering
		\begin{tabular}{@{}c|cccc@{}}
			\toprule
			\# Variations & PSNR   $\uparrow$        & SSIM    $\uparrow$        & LPIPS    $\downarrow$   & rendering time (s)    $\downarrow$      \\ \midrule
			(a)          & 34.03          & .9745          & .0404  &  .1237       \\
			(b)        & 34.61          & .9822          & .0263 & .1529     \\
			(c)         & 34.70          & .9829          & .0257 & .1847          \\
			(d)         & 35.82 & .9831 & .0249 & .2331\\ \bottomrule
		\end{tabular}
		\caption{Quantitative evaluation on the number of the sample points in terms of various metrics. (a) using two residual blocks and one 'Upconv' layer ; (b) ours (three residual block); (c) using six residual blocks; (d) using nine residual blocks.
		}
		\label{table:feature}
		\vspace{-3mm}
		\vspace{-10pt}
	\end{table}
	
	\vspace{1mm}\noindent{\bf Ablation study of refinement network.}
	We provide qualitative evaluation of our refinement network in Fig. \ref{eval1}.

\end{document}